\documentclass[3p,times,singlecolumn]{elsarticle}
\usepackage[utf8]{inputenc}
\usepackage{lineno,hyperref} 
\usepackage{graphicx,svg,subcaption,placeins}
\usepackage{colortbl,multirow,booktabs} 
\usepackage{setspace} 
\usepackage{xcolor} 
\biboptions{compress}

\journal{Knowledge-Based Systems}
\date{December 22, 2022}

\begin{document}

	\begin{frontmatter}
		
		\title{STB-VMM: Swin Transformer Based Video Motion Magnification} 
		
		\author[IQS]{Ricard Lado-Roigé}
		\author[IQS]{Marco A.  Pérez\corref{cor1}}
		\ead{marcoantonio.perez@iqs.url.edu}
		\cortext[cor1]{Corresponding author.}
		\address[IQS]{IQS School of Engineering, Universitat Ramon Llull, Via Augusta 390, 08017 Barcelona, Spain}
		
		\begin{abstract}
			The goal of video motion magnification techniques is to magnify small motions in a video to reveal previously invisible or unseen movement. Its uses extend from bio-medical applications and deepfake detection to structural modal analysis and predictive maintenance. However, discerning small motion from noise is a complex task, especially when attempting to magnify very subtle, often sub-pixel movement. As a result, motion magnification techniques generally suffer from noisy and blurry outputs. This work presents a new state-of-the-art model based on the Swin Transformer, which offers better tolerance to noisy inputs as well as higher-quality outputs that exhibit less noise, blurriness, and artifacts than prior-art. Improvements in output image quality will enable more precise measurements for any application reliant on magnified video sequences, and may enable further development of video motion magnification techniques in new technical fields.
		\end{abstract}
		
		
	\end{frontmatter}

	{\color{black}\section{Introduction}		
		Video Motion Magnification (VMM) is a computer vision task consistent in magnifying small motions in a video sequence, having several uses in many fields from bio-medical applications \cite{bio_ref, mcleod2014motion, lauridsen2019extracting} and deepfake detection \cite{fei2021exposing} to structural modal analysis \cite{Lado2022} and condition monitoring. These techniques act like a microscope for motion, to reveal previously invisible or unseen movements. Despite this simple premise, discerning small motions from noise is a complex task, especially when attempting to magnify very subtle, often sub-pixel movement. As a result, motion magnification techniques generally suffer from noisy and blurry outputs. Therefore, multiple authors have explored techniques to remediate these shortcomings and improve magnification quality and performance. 
		
		Early motion magnification algorithms, such as \cite{liu_motion_2005}, used a Lagrangian approach, reliant on motion tracking or optical flow, to isolate motion prior to magnification. However, this approach is very computationally expensive and difficult to execute artifact-free, especially in regions affected by occlusion boundaries and complex motion. On the other hand, more modern techniques \cite{wu_eulerian_2012, wadhwa_phase-based_2013,  wadhwa_riesz_2014, wadhwa_eulerian_2016, zhang_video_2017} have relied on Eulerian approaches, which observe the changes in a fixed region of pixels instead of tracking features in time and space. These Eulerian approaches are less computationally expensive, perform better with small motions, and generally yield better magnification results. Nevertheless, these approaches still display noticeable blurring and artifacting due to the complex challenge of designing filters for noise removal, which at the same time, do not interfere with motion magnification. For this reason, Oh et al. \cite{oh_learning-based_2018} proposed a novel learning-based approach to VMM. Learning-based motion magnification departs from the use of hand-designed filters in favor of learning those filters using Convolutional Neural Networks (CNN) instead. This method achieved higher-quality magnification yielding fewer ringing artifacts and showing better noise characteristics than previously published methods. However, its reliance on additional temporal filtering to improve image quality sometimes produces errors in magnification. While it is possible to obtain fairly clear results with no temporal filtering, the image quality generally improves when filtering is applied as it removes unwanted motion and noise before learning-based magnification.
		
		The method presented in this work improves on the learnable filters and abandons temporal filtering to ensure correct magnification outputs. Resulting in a novel architecture capable of producing state-of-the-art results in terms of magnified image quality. The main contributions of this work are: 
		\begin{enumerate}[a)]
			\item A novel motion magnification architecture based on the SWIN transformer.
			\item A discussion, comparison, and validation of learning-based VMM techniques, both in a quantitative and qualitative sense.
			\item The proposed novel architecture outperforms relevant VMM techniques in both quantitative evaluation and observed output quality, offering higher-quality magnification, less blurry frame reconstruction, better noise tolerance, and fewer artifacts than prior-art.
		\end{enumerate}
		
		The following section summarizes previous influential works and their relation to the development of the presented model. Section three describes in detail the model's architecture and its training process. The fourth section presents results and comparisons of the model's performance, focusing on magnification and image quality with respect to prior work. Finally, the conclusions of this paper are summarized in section five.
		}
	
	\section{Related work}		
		\subsection{Learning-based video motion magnification}
			Eulerian approaches to video motion magnification function by decomposing video sequences into motion representations that can later be manipulated mathematically and then reconstructed into magnified frames. On the other hand, Lagrangian approaches explicitly track a pixel or feature's movement throughout a video sequence. This distinction between Lagrangian and Eulerian approaches is not dissimilar to the same terms used in fluid dynamics, where Lagrangian methods \cite{liu_motion_2005} track a volume of fluid through the flow, while Eulerian approaches \cite{wadhwa_phase-based_2013,wu_eulerian_2012,wadhwa_riesz_2014} study the evolution of flow in a fixed volume in space. Eulerian-based methods generally have the upper hand when processing small motion but produce blurry results when encountering large motion. The technique presented in this paper belongs to the Eulerian approach and is inspired by the work of Oh et al.'s learning-based video motion magnification \cite{oh_learning-based_2018}.
			
			\textcolor{black}{
				Eulerian techniques generally consist of three stages: spatial decomposition, motion isolation and manipulation, and representation denoising. From this blueprint, different authors have proposed increasingly sophisticated techniques to improve magnification quality and performance as reflected in table \ref{tab:PriorArts}. In technical terms, the motion magnification problem can be summarized as follows. Given a signal $I(x,t) $ representing image intensity at position $x$ and time $t$, and $\delta(t)$ representing translational motion in time such that
				\begin{equation}
					I(x,t)=f(x + \delta(t)); \ 
					I(x,0)=f(x)
				\end{equation}
				The goal of motion magnification is to synthesize the signal
				\begin{equation}
					\hat{I}(x,t)=f(x+(1+\alpha) \cdot \delta(t))
				\end{equation}
				for some amplification factor $\alpha$. In practice, only certain frequencies of motion $\delta(t)$ are useful to motion magnification, so a selector $T(\cdot)$  is applied to $\delta(t)$, which is typically a temporal bandpass filter.
			}
			
			\begin{table}[h]
				\centering
				\resizebox{1.00\linewidth}{!}{%
					{\color{black}\begin{tabular}{@{}p{2.2cm}p{2.5cm}p{2.5cm}p{2.5cm}p{2.5cm}p{2.5cm}p{2.5cm}p{2.5cm}@{}}
						\toprule
						\textbf{Method} &
						\textbf{Liu et al. \cite{liu_motion_2005}} &
						\textbf{Wu et al. \cite{wu_eulerian_2012}} &
						\textbf{Wadhwa et al. \cite{wadhwa_phase-based_2013}} &
						\textbf{Wadhwa et al. \cite{wadhwa_riesz_2014}} &
						\textbf{Zhang et al. \cite{zhang_video_2017}} &
						\textbf{LB-VMM \cite{oh_learning-based_2018}} &
						\textbf{STB-VMM} \\ \midrule
						\textbf{Spatial decomposition} &
						Tracking, optical flow &
						Laplacian pyramid &
						Steerable filters &
						Riesz pyramid &
						Steerable filters &
						Deep convolution layers &
						Swin Transformer \\ [30pt]
						\textbf{Motion isolation} &
						- &
						Temporal bandpass filter &
						Temporal bandpass filter &
						Temporal bandpass &
						Temporal bandpass filter (2$_{nd}$ order derivative) &
						Subtraction or bandpass filter &
						Subtraction \\ [30pt]
						\textbf{Representation denoising} &
						Expectation-Maximization &
						- &
						Amplitude weighted Gaussian filtering &
						Amplitude weighted Gaussian filtering &
						Amplitude weighted Gaussian filtering &
						Trainable convolution &
						Swin Transformer \\ \bottomrule
					\end{tabular}}%
				}
				\caption{Motion magnification techniques summary table. Adapted from \cite{oh_learning-based_2018}.}
				\label{tab:PriorArts}
			\end{table}
			
			Prior to learning-based VMM (LB-VMM), magnification techniques relied on multi-frame temporal filtering to isolate motions of interest from random noise \cite{wu_eulerian_2012, wadhwa_phase-based_2013,wadhwa_riesz_2014,Elgharib2015VideoMI, zhang_video_2017}. By contrast, the learning-based approach \cite{oh_learning-based_2018} directly employs CNNs to both filter noise and extract features, achieving comparable or better quality than prior-art without using temporal filtering. The LB-VMM model is composed of three stages: encoder, manipulator, and decoder. Said model is designed to accept two frames and return a single motion-magnified frame. The goal of the encoder is to extract relevant features from each of the two input frames and yield a visual and a motion representation. The motion representation of both input frames is then passed to the manipulator, which will subtract both representations and magnify the result by an arbitrary parameter $\alpha$ defined by the user. Finally, the results of the manipulator and the previously-obtained visual representation enter the decoder, where the motion and visual components are reconstructed into a motion-magnified frame. These three CNN-based components allow for flexible learnable filters that are better suited to the task of motion magnification and thus yield better quality magnification results.
			
			To train the model and given the impossibility of obtaining motion magnified video pairs, Oh et al. generated and used a fully-synthetic dataset for training their model, built by moving segmented objects from the PASCAL VOC \cite{everingham_pascal_2010} dataset over background images taken from MS COCO \cite{lin_microsoft_2015}. Careful consideration to the generation of the dataset was paid to ensure accurate pixel and sub-pixel motion as well as learnability. The dataset learning examples are parametrized to make sure they are within a defined range. Specifically, the dataset's magnification is upper-limited to an $\alpha$ magnification factor of 100, and input motion is sampled so that magnified motion does not exceed 30 pixels.
		
		{\color{black}\subsection{Transformers as a Computer Vision tool}
			CNNs have been a staple of the Computer Vision (CV) field in the last few years, with many of the top-performing models having made extensive use of them \cite{lecun1989,alexnet,He_residual}. This period roughly started after Krizhevsky et al. \cite{alexnet} won the ImageNet Large Scale Visual Recognition Challenge \cite{deng2009imagenet, ImageNetChallenge} (ILSVRC) on September 30$^{th}$ 2012, and spurred many publications employing CNNs and GPUs to accelerate deep learning. Through the use of filters, these networks generate feature maps that summarize an image's most relevant parts. These filters capture relevant local information by the very nature of the convolution operation, which, combined with multi-scale architectures \cite{unet, hrnet} result in rich feature maps that can efficiently obtain a representation of an image's content, both in a local and global context. Recently, the CV field has been revolutionized yet again by the Vision Transformer (ViT) \cite{ViT}, which, employing the attention mechanism has demonstrated state-of-the-art performance in many CV tasks. The attention mechanism was first popularized in the field of Natural Language Processing (NLP) by Vaswani et al. \cite{Vaswani2017AttentionIA}, where the transformer architecture has become the de-facto standard. 
			
			The attention mechanism can be described as mapping from a query and a set of key-value pairs into an output. The output, represented in vector format, is computed as a weighted sum of the values, where the weight assigned to each value is computed by a compatibility function taking into account the query and the corresponding key \cite{Vaswani2017AttentionIA}. The transformer was the first model which exclusively relied on self-attention to compute representations of its input and output without using sequence-aligned recursive neural networks or convolution operations. Unlike CNNs, transformers lack translation invariance and a locally-restricted receptive field, in its place transformers offer permutation invariance. Said feature enabled NLP models to infer relations between words and ideas much further into a text than previous recurrent models could. However, CV applications require the processing of grid-structured data which can not trivially be processed by a transformer. The ViT \cite{ViT} overcame this burden by mapping grid-structured data into sequential data by splitting the image into patches. Patches are then flattened into vectors and embedded into a lower dimension. These flattened patches are then summed with positional embeddings and fed as a sequence to a standard transformer encoder. Image patches essentially become sequence tokens just like words are when working in NLP, in fact, ViT uses the exact same encoder described in \cite{Vaswani2017AttentionIA}.
			
			Later, Microsoft researchers improved on the ViT publishing the SWIN transformer, a hierarchical vision transformer using shifted windows \cite{SWIN}. This work further refined the solution to adapt the original transformer from language to vision. The SWIN transformer solved issues caused by large discrepancies in the scale of visual entities at the same time that limited self-attention computation to non-overlapping local windows, yet still allowing for cross-window interaction. The introduced limitation on the scope of self-attention significantly reduced the computational complexity, which scales quadratically with respect to image size, allowing for the processing of higher-resolution images that were previously unmanageable. Further developments in the CV field have implemented the SWIN transformer for various tasks achieving state-of-the-art performance \cite{TransformersSurvey, SWINIR, VRT}.
		}
			
		\subsection{SwinIR image restoration}
			Inspired by the recent prominence of the transformer and its success in many CV problems such as image classification \cite{Ramachandran_2019,ViT, bichen_2020, SWIN, Yawei_2021,Liu_2021,Vaswani_2021}, object detection \cite{Carion_2020, liu_deep_2020, Touvron_2020}, segmentation \cite{bichen_2020,Zheng_2020,Cao_2021_unet}, crowd counting \cite{liang_transcrowd_2022, Guolei_2021} and image restoration \cite{Chen_2020, Cao_2021_videoSR, Wang_2021}, Liang et al. \cite{SWINIR} proposed a new state-of-the-art image restoration model based on the Swin transformer \cite{SWIN}. The SwinIR model consists yet again of three modules: a shallow feature extractor, a transformer-based deep feature extractor and a high-quality image reconstruction module. This structure offers excellent performance in various image restoration tasks such as image super-resolution, JPEG compression artifact reduction, and image denoising. These applications are very interesting when working with VMM, as current state-of-the-art methods can be negatively affected by noisy input images, causing much noisier and blurrier results, especially at large magnification rates. This occurs as a result of noise not being properly filtered beforehand, therefore as the motion gets magnified, the noise gets magnified as well.
		
	\section{Methodology}
					{\color{black}\subsection{Residual Swin Transformer Block}
				 	The Residual Swin Transformer Block (RSTB) \cite{SWINIR} is used as one of the fundamental building blocks of the proposed architecture, appearing in parts of both the feature extractor and the reconstructor. The RSTB is a residual block combining multiple Swin Transformer Layers (STL) \cite{SWIN} and convolutional layers, compounding the benefits of the spatially invariant filters of the convolutional layers with the residual connections that allow for multilevel feature processing.
					
					The Swin transformer layer shown in figure \ref{fig:arch_details} partitions an $H \times W \times C$ image into non-overlapping $\frac{HW}{M^2}$ local windows using an $MxM$ sliding window and then computing its local attention, effectively reshaping the input image into $\frac{HW}{M^2} \times M^2 \times C$. The main difference with respect to the original transformer layer \cite{Vaswani2017AttentionIA} lies in the local attention and the shifted window mechanism. For a local window feature $F \in \mathbb{R}^{M^2 \times C}$, the query, key, and value matrices $Q$, $K$, and $V$ $\in \mathbb{R}^{M^2 \times d}$ are computed as
					
					\begin{equation} \label{QKV}
						Q = F W_Q;\quad K = F W_k;\quad V = F W_V
					\end{equation}
					where $W_Q$, $W_K$, and $W_V$ are the learnable parameters shared across different windows, and $d$ is the dimension of $Q$, $K$, and $V$. Therefore, the attention matrix is computed for each window as
					\begin{equation} \label{Attention}
						Attention(Q,K,V) = softmax(\frac{Q K^T}{\sqrt{d}}+P) V
					\end{equation}
					where $P$ is the learnable relative positional encoding. Computing the attention mechanism multiple times yields the results of the Multi-head Self Attention (MSA), which are then passed on to a Multi-Layer Perceptron (MLP). Therefore, the whole STL process can be summed up like so
					\begin{equation} \label{STL1}
						F = MSA(LayerNorm(F)) + F
					\end{equation}
					then
					\begin{equation} \label{STL2}
						F = MLP(LayerNorm(F)) + F
					\end{equation}
					where the MLP is formed by two fully-connected layers with a GELU activation layer in between.}
					
		{\color{black}\subsection{Network architecture}
			The proposed model architecture, shown in figure \ref{fig:arch_pic}, consists of three main functional blocks: the feature extractor, the manipulator, and the reconstructor. The feature extractor is further subdivided into the shallow and deep feature extractors, and their job is to extract a high-quality representation of an input frame. Next, the manipulator, using the features from two frames, magnifies the motion by multiplying the difference between the two feature spaces by a user-selected magnification factor $\alpha$. Finally, the reconstructor converts the resulting manipulated feature space back into a magnified frame. 
			
			\begin{figure}[h]
				\centering
				\includegraphics[width=0.60\textwidth]{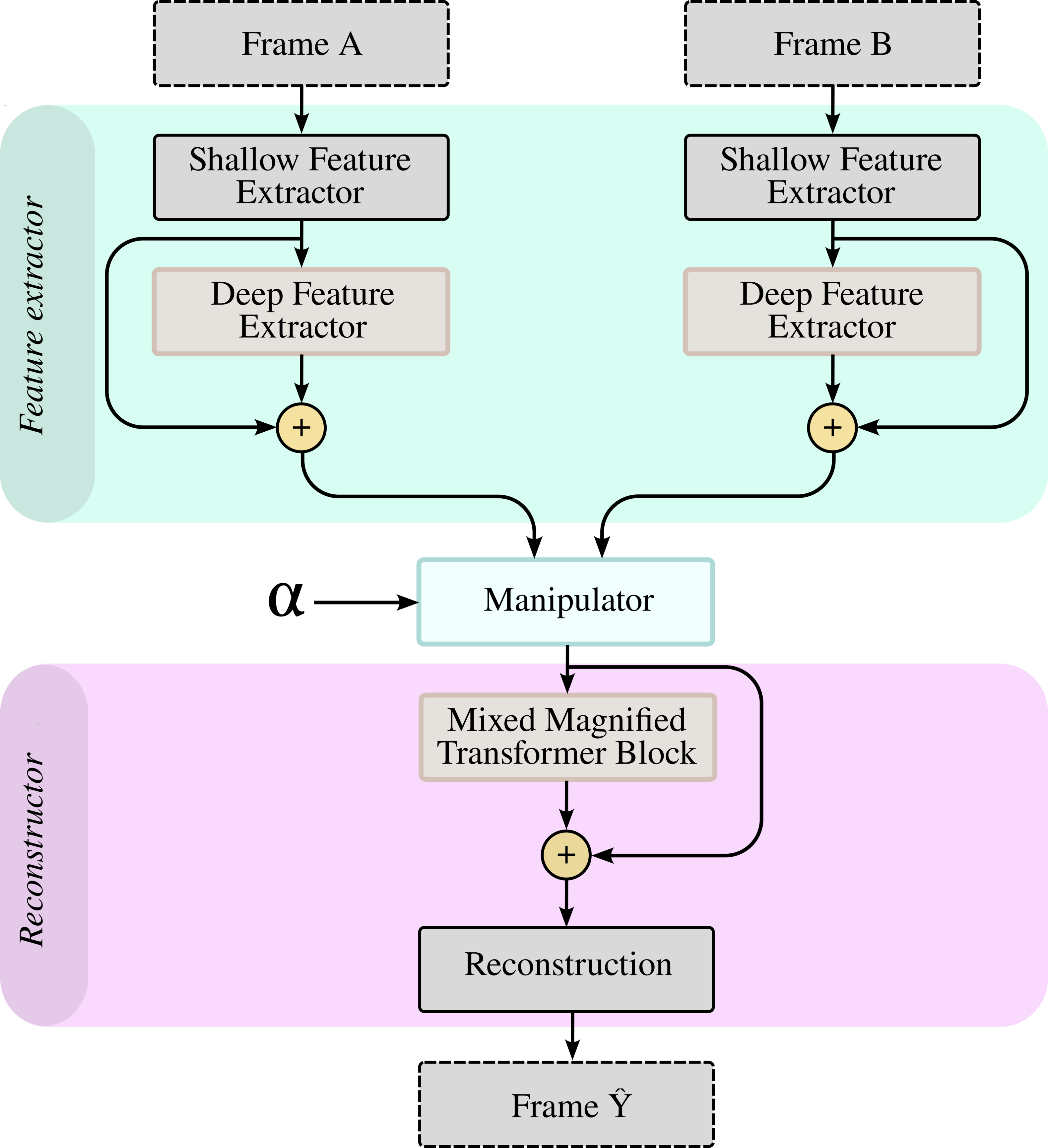}
				\caption{Architecture overview of the proposed model.}
				\label{fig:arch_pic}
			\end{figure}
			
			Given two frames of a target sequence $[I_A, I_B] \in \mathbb{R} ^{H \times W \times C_{in}}$ (where $H$ is the height of the image, $W$ is the width of the image and $C_{in}$ represents the number of input channels) the convolutional shallow feature extractor ($G_{SF}$) maps high-level features into a higher dimensional feature space, thus providing early local feature extraction ($F_{AS}, F_{BS}$) and leading to a more stable optimization and better results \cite{xiao_2021}.
			
			\begin{equation} \label{Gsf}
				[F_{AS}, F_{BS}] = G_{SF}([I_A, I_B])
			\end{equation}
			
			Then, the features extracted in the previous step are further processed in the deep feature extraction module ($G_{DF}$), which consists of $N$ Residual Swin Transformer Blocks (RSTB).
			
			\begin{equation} \label{Gdf}
				[F_{AD}, F_{BD}] = G_{DF}([F_{AS}, F_{BS}])
			\end{equation}
						
			After feature extraction, both frames' feature spaces are then sent to the manipulator \cite{oh_learning-based_2018} ($G_M$), which works by taking the difference of both frames' feature spaces and directly multiplying by a magnification factor $\alpha$.
			
			\begin{equation} \label{Gm1}
				G_{M}(F_{AS} + F_{AD}, F_{BS} + F_{BD}) = (F_{AS} + F_{AD}) + h(\alpha \cdot t(((F_{BS} + F_{BD})-(F_{AS} + F_{AD})))) 
			\end{equation}
			
			Where $t(\cdot)$ is a $3 \times 3$ convolution followed by a ReLU activation, and $h(\cdot)$ is a $3 \times 3$ convolution followed by a $3 \times 3$ residual block.
			
			\begin{equation} \label{Gm2}
				F_{M} = G_{M}(F_{AS} + F_{AD}, F_{BS} + F_{BD})
			\end{equation}
			
			The conjoined manipulated feature space of both frames is then processed by the Mixed Magnified Transformer Block (MMTB) ($G_{MMTB}$) formed by $N$ RSTB blocks. This stage enables the attention mechanism to affect the combined magnified features of both frames, resulting in a more coherent result after reconstruction.
			
			\begin{equation} \label{Gmmtb}
				F_{MMTB} = G_{MMTB}(F_{M})
			\end{equation}
			
			Finally, reconstruction is dealt with a convolutional block ($G_R$) that inverts the initial feature mapping, done in the shallow feature extractor, back onto a frame ($I_{\hat{Y}}$).
			
			\begin{equation} \label{Gr}
				I_{\hat{Y}} = G_{R}(F_{M} + F_{MMTB})
			\end{equation}
		
			Further detail on the architecture can be found in figure \ref{fig:arch_details} along with a graphical representation of the Swin Transformer Layer (STL) and the Residual Swin Transformer Block (RSTB).
			
			\begin{figure}[h]
				\centering
				\begin{subfigure}[b]{0.4\linewidth}
					\caption{Swin Transformer Layer (STL)}
					\includegraphics[width=\linewidth]{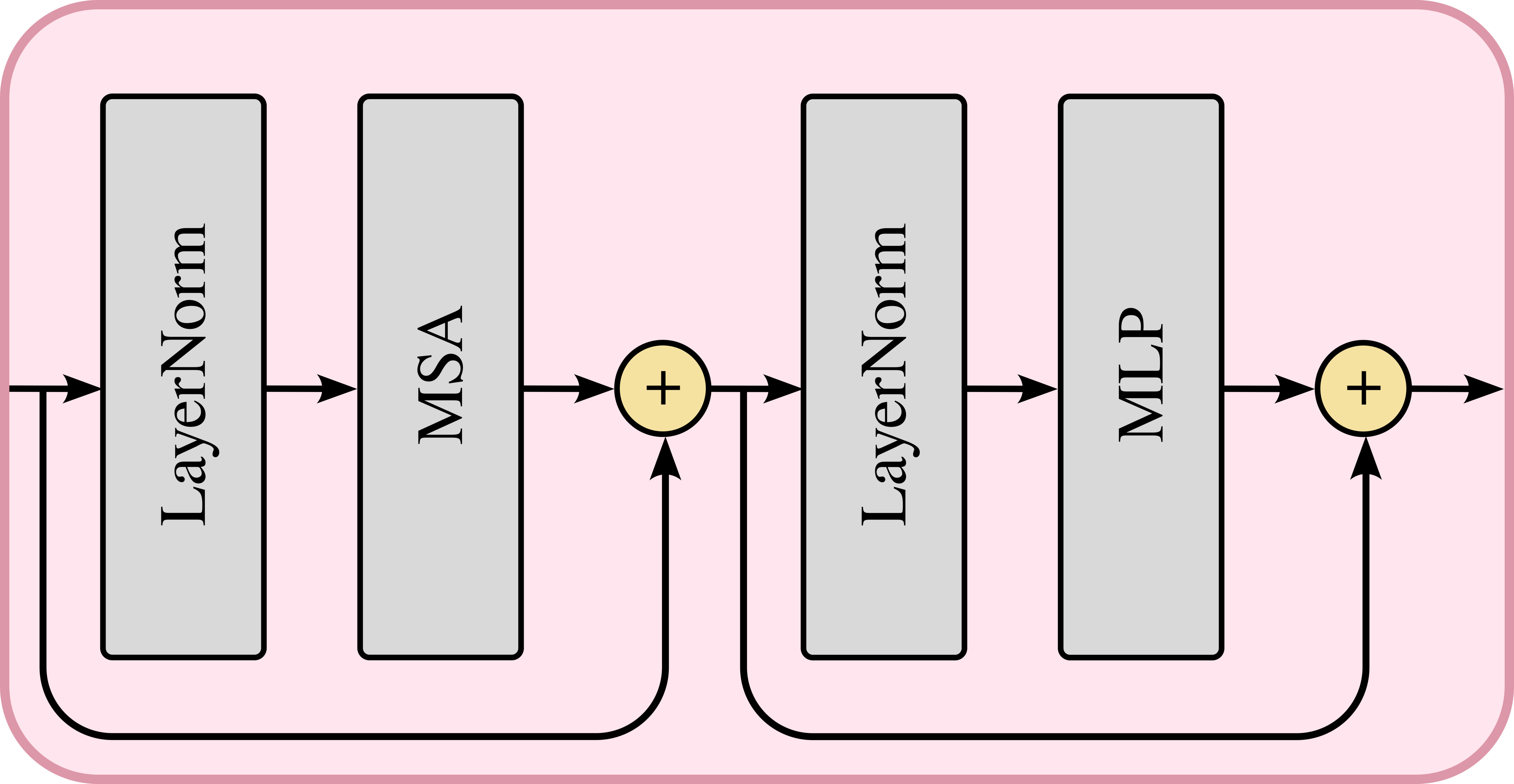}
				\end{subfigure}
				\hspace{0.05\linewidth}
				\begin{subfigure}[b]{0.21\linewidth}
					\caption{Residual Swin Transformer Block (RSTB)}
					\includegraphics[width=\linewidth]{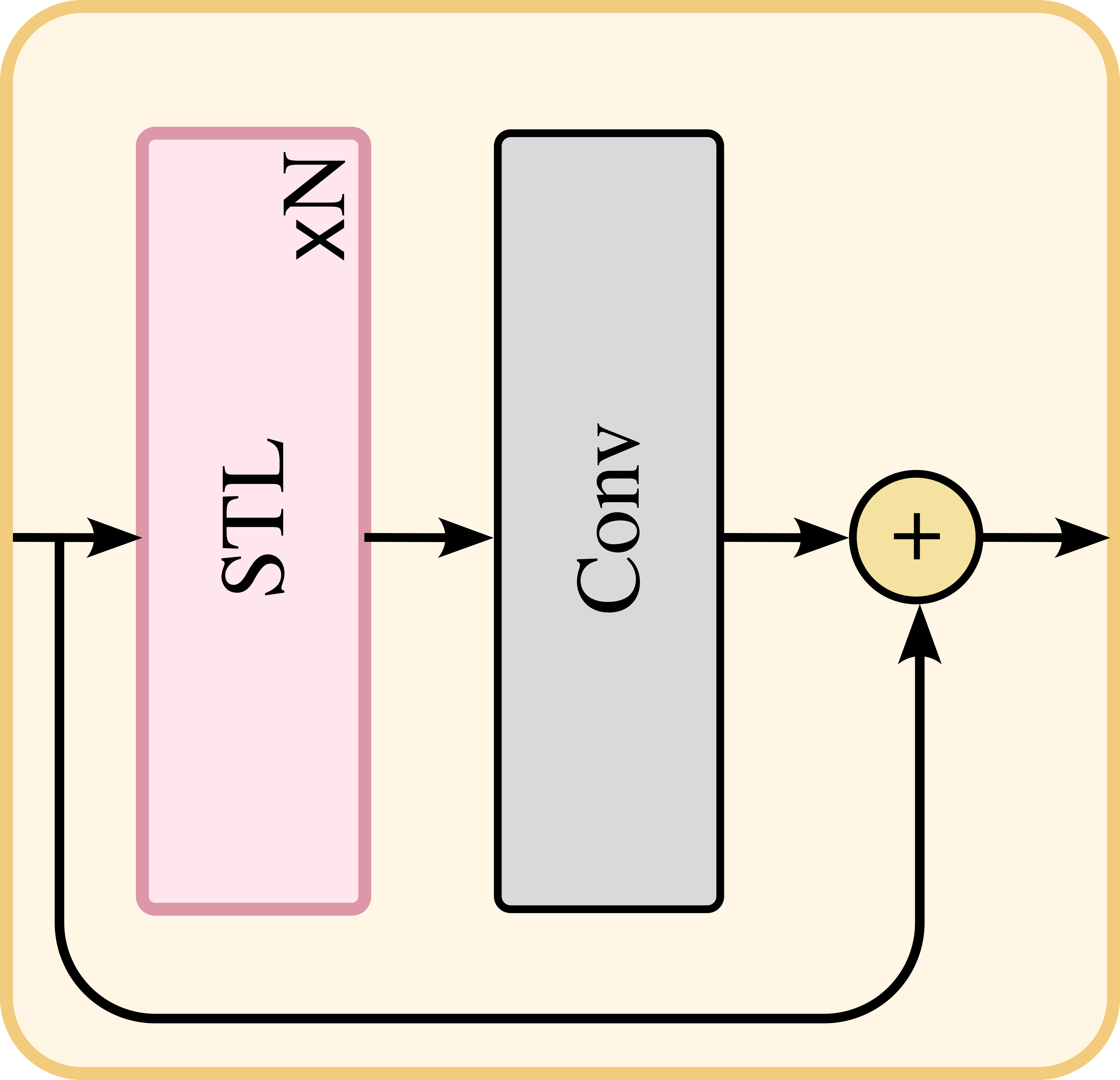}
				\end{subfigure}
				\hspace{0.05\linewidth}
				\begin{subfigure}[b]{0.21\linewidth}
					\caption{Deep Feature Extractor / Mix Magnified Transformer Block}
					\includegraphics[width=\linewidth]{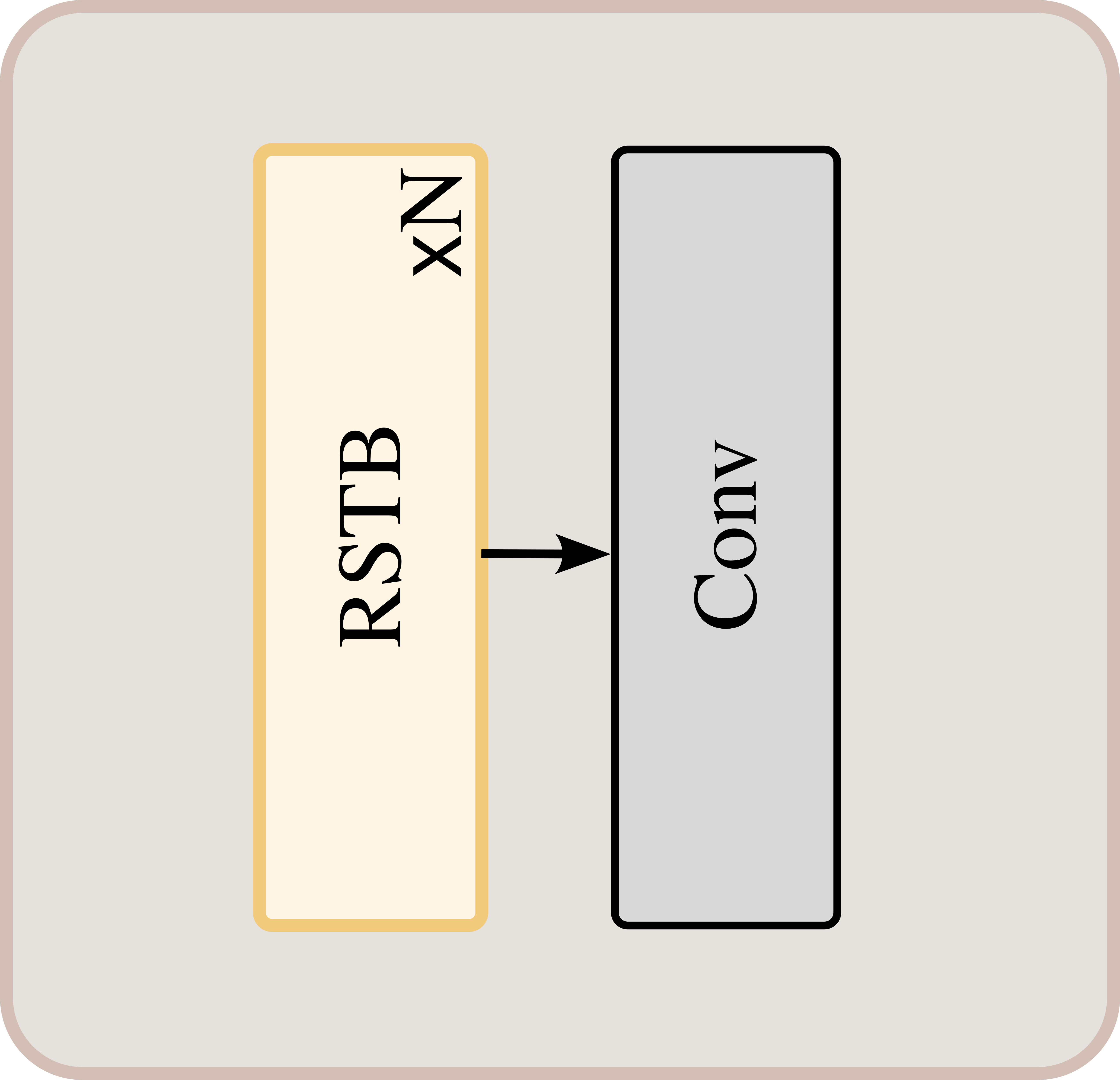}
				\end{subfigure}
				\\
				\vspace{10pt}
				\begin{subfigure}[b]{0.40\linewidth}
					\caption{Manipulator}
					\includegraphics[width=\linewidth]{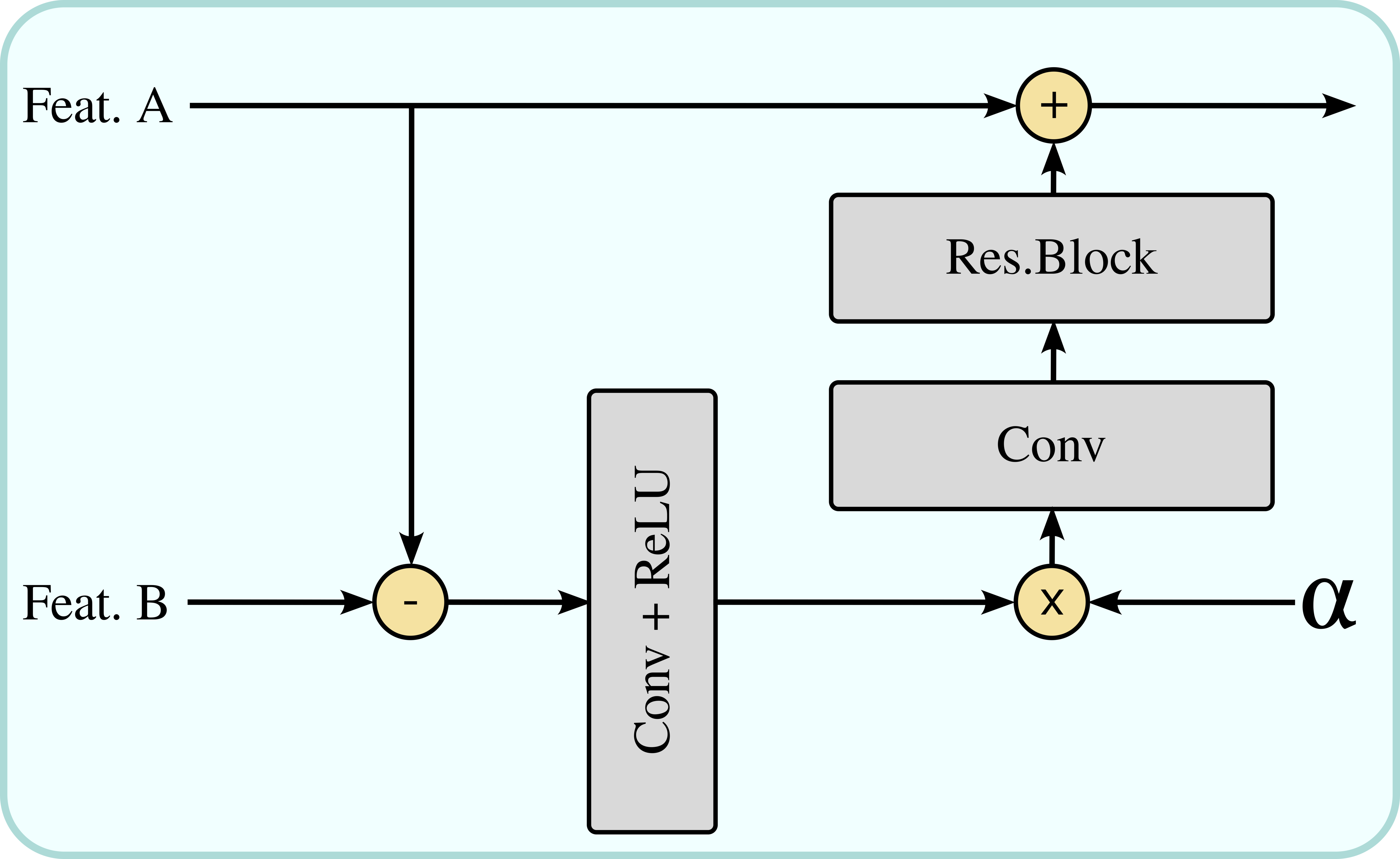}
				\end{subfigure}
				
				\caption{Architectural details.}
				\label{fig:arch_details}
			\end{figure}
		}
		
		\subsection{Training}
			The whole network is trained end-to-end using the dataset provided by \cite{oh_learning-based_2018}, which allows the results comparison to depend exclusively on network architecture. Nevertheless, in addition to enabling fair comparison, the dataset has proven \cite{oh_learning-based_2018} to produce good-quality models capable of generalizing trained scenarios and returning excellent-quality magnified videos on scenes totally unrelated to the dataset. These reasons led to the adoption of the dataset as the only source of training data.
			
			The L1-Loss cost function was chosen for end-to-end training and placed between the network's output $I_{\hat{Y}}$ and the ground truth frame $I_{Y}$. Additionally, in order to improve the feature extraction and make a more robust system, the perturbed $c$ frames provided by the dataset were compared against their non-perturbed counterparts after feature extraction, using yet again L1-Loss. The resulting regularization loss was then added to the end-to-end loss of the whole network with a $\lambda$ weight coefficient set to 0.1.
			
			Finally, the optimizer of choice for training the model was ADAM \cite{Kingma2015AdamAM} with $\beta_1=0.9$, $\beta_2=0.999$, batch size set to 5 and a learning rate of $10^{-5}$ with no weight decay.
		
		\subsection{Modes of operation}
			The proposed approach, STB-VMM, can be applied to any input video sequence containing two frames or more, regardless of the time scale between the two frames. Sequences can be treated in one of two modes, static or dynamic, borrowed from \cite{oh_learning-based_2018}. No changes to the network are made for these modes. Instead, the modes refer to the order in which the input frames are fed to the model. The static mode, which follows more closely the classical definition of motion magnification, uses the first frame of the sequence as reference. In terms of computation, the static mode can be expressed like so: $model(I_0, I_t)$, where the $t$ is the frame number increasing sequentially with time. On the other hand, the dynamic mode magnifies the difference between two consecutive frames [$model(I_t, I_{t+1})$], therefore magnifying velocity between each frame. Note that in each of the modes, the magnification factor $\alpha$ has different meanings.
			
			Oh et al. \cite{oh_learning-based_2018} proposed one additional operation mode with temporal filtering to mitigate the effects of undesired motion and noise. The filtering was applied in the manipulator to produce temporarily-filtered motion-magnified frames similar to those of classical techniques. On the downside, the temporal mode appears to cause blindness to small motions, resulting in patchy magnification. This phenomenon occurs because motion amplitude crosses the threshold to be large enough to be detected and causes some regions to be suddenly magnified mid-sequence. This performance degradation gets worst when the magnification factor is high and motion is small. While theoretically possible to incorporate a temporal mode into the proposed model, the magnification results do not suffer from excessive noise or blurring, therefore, temporal filtering is unnecessary and the full spectrum of frequencies is magnified all at once producing good results.
		
	\section{Results and discussion}
		In the following section, the results yielded by the STB-VMM model are compared to the current state-of-the-art learning-based video motion magnification model \cite{oh_learning-based_2018}. Performance is measured quantitatively and qualitatively, showing that our model improves on the previous state-of-the-art in magnification quality and clarity. The video versions of all the comparisons are available in the supplementary materials.
		
		Quantitative comparison of image quality or Image Quality Assessment (IQA) is a complex topic involving many variables and methods. Said methods are divided into three main categories: full-reference, reduced-reference, and no-reference. A referenced algorithm \cite{psnr_Onur_2021, SSIM_Zhou_2004, Lanjiang_2021_IQA} requires a pristine sample to assess the quality of a degraded image, while no-reference methods \cite{Mittal2012NoReferenceIQ, Mittal_2013, venkatanath2015blind, Junjie_2021_MUSIQ} produce an image score without the need of any reference. When evaluating VMM it is impossible to obtain a pristine motion-magnified frame. Therefore, to evaluate the results presented in the following section the MUSIQ \cite{Junjie_2021_MUSIQ, pyIQA} algorithm was chosen to compare the models' performance.
		
		The following results comparative analyzes the performance of Oh et al.'s Learning-Based Video Motion Magnification (LB-VMM) model and STB-VMM on ten different video benchmarks that showcase interesting motion magnification examples. In addition, a comparison against the baby \cite{wu_eulerian_2012} sequence is added to provide a fair point of comparison. The sequences were captured at 1080p 60fps on a mid-range smartphone to demonstrate the potential of STB-VMM with accessible video equipment.
		
		\subsection{Quantitative comparison}
			Table \ref{tab:Benchmark} shows the average, 1st, and 99th percentile average MUSIQ scores for the tested benchmark sequences ran on the Learning-based Video Motion Magnification model and the STB-VMM model. The values presented in the table are calculated for each individual frame of the full sequences and then summarized on an average score. The original sequences are also added as control, and scores are expected to be higher than both of the magnification methods.
			
			\begin{table}[h]
				\centering
				\resizebox{0.85\linewidth}{!}{%
					\begin{tabular}{llllllllll}
						&
						\multicolumn{3}{l}{\textbf{Original}} &
						\multicolumn{3}{l}{\textbf{LB-VMM}} &
						\multicolumn{3}{l}{\textbf{STB-VMM}} \\ \hline
						\textbf{} &
						Avg.  &
						$\eta_{1}$  &
						$\eta_{99}$ &
						Avg.  &
						$\eta_{1}$  &
						$\eta_{99}$ &
						Avg.  &
						$\eta_{1}$  &
						$\eta_{99}$ \\ 
						\hline
						AC$_{00}$       & 72.11 & 69.65 & 72.75 & 55.73 & 49.61 & 58.69 & 62.45 & 61.05 & 63.29 \\
						AC$_{01}$        & 69.15 & 68.30 & 70.05 & 48.35 & 34.07 & 51.22 & 59.27 & 57.72 & 60.96 \\
						Baby         & 74.39 & 69.71 & 74.87 & 55.51 & 53.26 & 59.95 & 57.12 & 54.41 & 62.90 \\
						Building$_{00}$  & 66.84 & 66.01 & 75.45 & 52.46 & 49.51 & 62.75 & 52.30 & 50.07 & 56.43 \\
						Car$_{00}$       & 52.55 & 50.65 & 54.41 & 31.40 & 18.27 & 35.50 & 43.37 & 23.28 & 48.06 \\
						Car$_{01}$       & 55.81 & 54.77 & 57.01 & 33.51 & 30.67 & 64.99 & 50.28 & 48.08 & 52.07 \\
						Crane$_{00}$     & 75.26 & 74.86 & 75.57 & 56.92 & 52.70 & 65.02 & 59.13 & 56.19 & 62.89 \\
						Crane$_{01}$     & 75.09 & 74.63 & 75.44 & 51.05 & 45.25 & 57.37 & 54.93 & 51.11 & 64.70 \\
						Truss$_{00}$     & 66.94 & 65.92 & 67.49 & 55.90 & 52.65 & 57.98 & 56.27 & 54.93 & 57.61 \\
						Wheel$_{00}$     & 72.84 & 71.87 & 73.38 & 51.04 & 28.82 & 54.40 & 57.04 & 36.41 & 61.19 \\
						Wheel$_{01}$     & 52.15 & 50.23 & 53.55 & 34.84 & 31.12 & 59.03 & 46.21 & 43.68 & 48.48 \\ \hline
						\textbf{Total avg.} &
						\textbf{66.13} &
						\textbf{51.25} &
						\textbf{75.45} &
						\textbf{48.05} &
						\textbf{32.32} &
						\textbf{60.09} &
						\textbf{54.42} &
						\textbf{45.68} &
						\textbf{63.29} \\ 
						\hline
						\textit{\textbf{\% dev. to avg.}} &
						\textit{13.58\%} &
						\textit{22.50\%} &
						\textit{14.09\%} &
						\textit{20.34\%} &
						\textit{32.75\%} &
						\textit{25.04\%} &
						\textit{10.70\%} &
						\textit{16.07\%} &
						\textit{16.28\%}
					\end{tabular}%
				}
				\caption{Comparative MUSIQ scores of the original sequence, the sequence magnified using Learning-Based Video Motion Magnification (\textit{o3f\_hmhm2\_bg\_qnoise\_mix4\_nl\_n\_t\_ds3} checkpoint), and the proposed method. (x20)}
				\label{tab:Benchmark}
			\end{table}
			
\begin{figure}[h]
				\centering
				\includegraphics[width=0.95\linewidth]{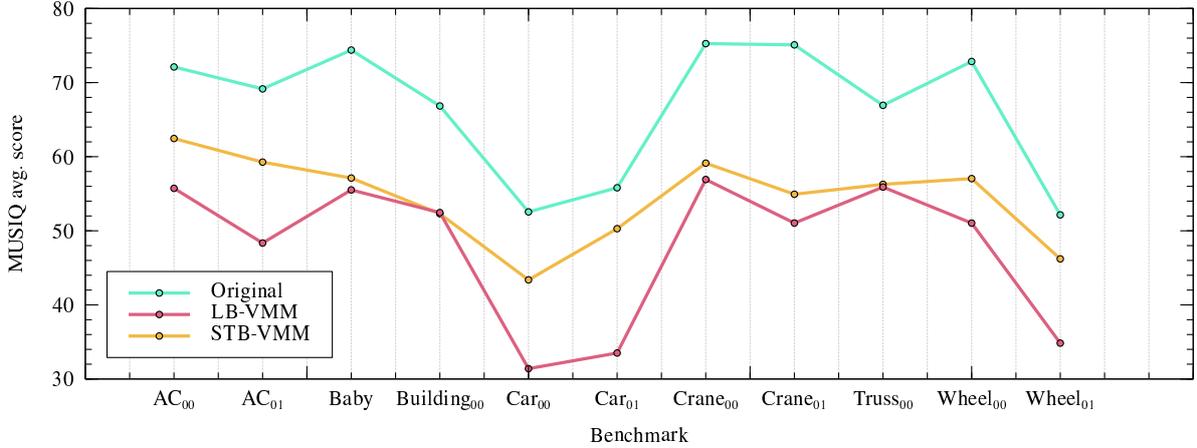}
				\caption{Graphic representation of the average MUSIQ scores per test sequence magnified x20.}
				\label{fig:Benchmark}
			\end{figure}

			The results on table \ref{tab:Benchmark} demonstrate that STB-VMM produces better results than LB-VMM. On average, the scores obtained by STB-VMM are 9.63\% higher, and  boast a much higher 1\% lows, implying that the quality of magnification is noticeably more consistent throughout the sequence.  This trend can be observed in table \ref{tab:Benchmark_diff} and figure \ref{fig:Benchmark}, where STB-VMM shows remarkable stability on its output quality. LB-VMM only manages a single higher score than STB-VMM in the building benchmark by a difference of 0.23\% (Building$_{00}$). However, in the authors' opinion, STB-VMM produces better quality magnification with more stable edges and less blurry patches. 
			
			On the other hand, none of the magnified scores fall above the original's, as expected. Nevertheless, magnified and original scores follow the same trend, implying that low-quality source videos produce worse outputs. However, STB-VMM is much more capable of dealing with low-quality input images, even closing the quality gap with respect to the original when input quality declines. The sharp quality declines seen in both car sequences can be, in part, attributed to the poor low light performance of the camera employed.
		
			\begin{table}[h]
				\centering
				\resizebox{0.375\linewidth}{!}{%
					\begin{tabular}{lrrr}
						& \textbf{Avg. (\%)} & \textbf{$\eta_{1}$ (\%)} & \textbf{$\eta_{99}$ (\%)} \\ \hline
						AC$_{00}$       & 9.32              & 16.42            & 6.32              \\
						AC$_{01}$       & 15.79             & 34.64            & 13.91             \\
						Baby        & 2.15              & 1.65             & 3.95              \\
						Building$_{00}$       & -0.23             & 0.86             & -8.38             \\
						Car$_{00}$       & 22.79             & 9.89             & 23.08             \\
						Car$_{01}$       & 30.06             & 31.78            & -22.65            \\
						Crane$_{00}$       & 2.93              & 4.67             & -2.81             \\
						Crane$_{01}$       & 5.17              & 7.85             & 9.71              \\
						Truss$_{00}$       & 0.55              & 3.46             & -0.55             \\
						Wheel$_{00}$       & 8.24              & 10.56            & 9.26              \\
						Wheel$_{01}$       & 21.81             & 25.01            & -19.72            \\ \hline
						\textbf{Total} & \textbf{9.63}     & \textbf{26.07}   & \textbf{4.24}    
					\end{tabular}%
				}
				\caption{MUSIQ score difference between STB-VMM and LB-VMM.}
				\label{tab:Benchmark_diff}
			\end{table}
		
		\subsection{Qualitative comparison}
			To reinforce the previous section's scores and claims, this section presents a few qualitative comparisons that demonstrate the effectiveness of our proposed network against the current state-of-the-art in terms of resulting image quality.
			
			\begin{figure}[h]
				\centering
				\begin{subfigure}[b]{0.909\linewidth}
					\includegraphics[width=\linewidth]{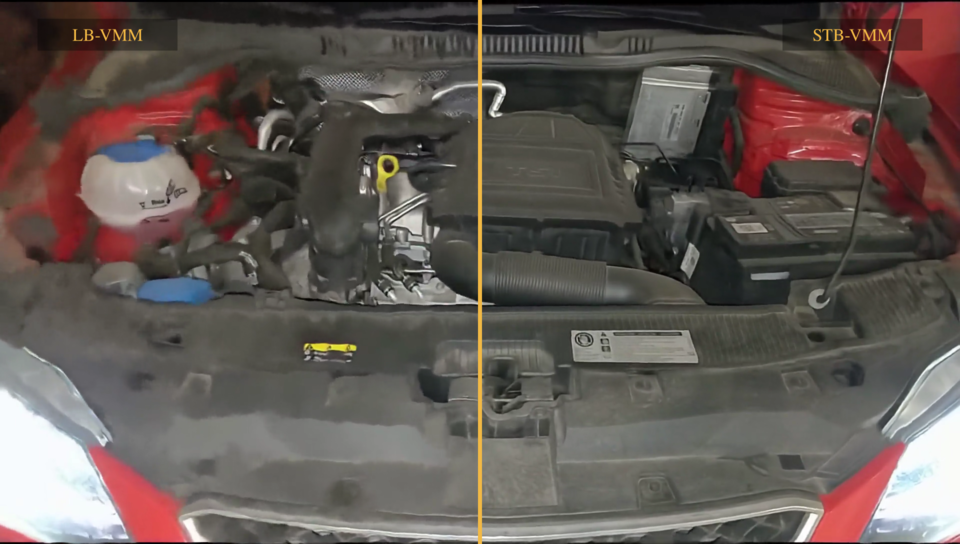}
					\caption{Split frame showing LB-VMM on the left and STB-VMM on the right}
				\end{subfigure}
				\begin{subfigure}[b]{0.242\linewidth}
					\includegraphics[width=\linewidth]{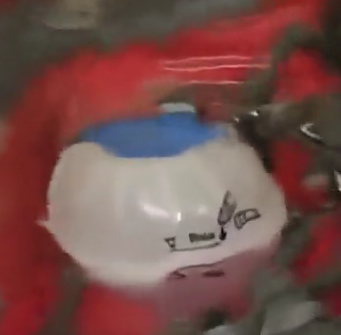}
					\caption{LB-VMM}
				\end{subfigure}
				\begin{subfigure}[b]{0.242\linewidth}
					\includegraphics[width=\linewidth]{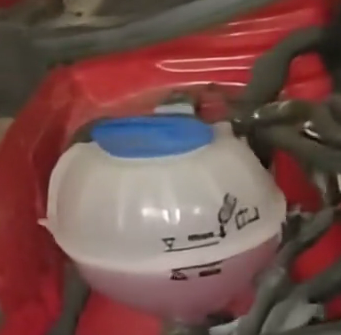}
					\caption{STB-VMM}
				\end{subfigure}
				\begin{subfigure}[b]{0.205\linewidth}
					\includegraphics[width=\linewidth]{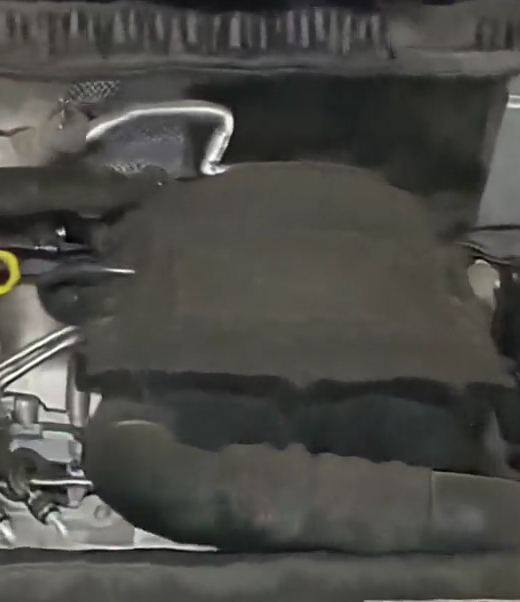}
					\caption{LB-VMM}
				\end{subfigure}
				\begin{subfigure}[b]{0.205\linewidth}
					\includegraphics[width=\linewidth]{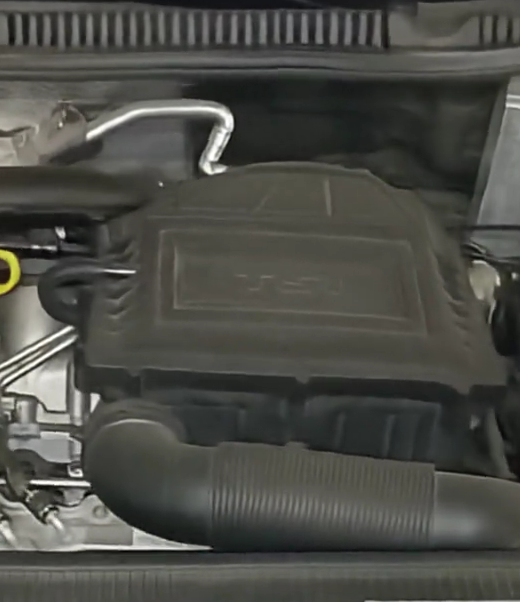}
					\caption{STB-VMM}
				\end{subfigure}
				
				\caption{Qualitative comparison of the car sequence. Highlighted in the bottom row of the figure the car's coolant reservoir, engine cover, and ventilation slits demonstrate that STB-VMM results are noticeably sharper and less distorted.}
				\label{fig:Car}
			\end{figure}
			
			Figure \ref{fig:Car} shows the same frame chosen at random from the Car$_{00}$ sequence using both models. STB-VMM, shown on the right, yields a much superior result in terms of image clarity that can be appreciated in both edges and texture. 
			
			The car sequence recording was filmed in a rather low light environment, thus yielding noisier/grainier video than otherwise could have been archived. This highlights one of the main benefits of the proposed architecture, which is a much better tolerance to noisy input. Regardless of clarity, both models perform well on motion magnification with very few artifacts, if any.
		
			\begin{figure}[h]
				\centering
				\begin{subfigure}[b]{0.9155\linewidth}
					\includegraphics[width=\linewidth]{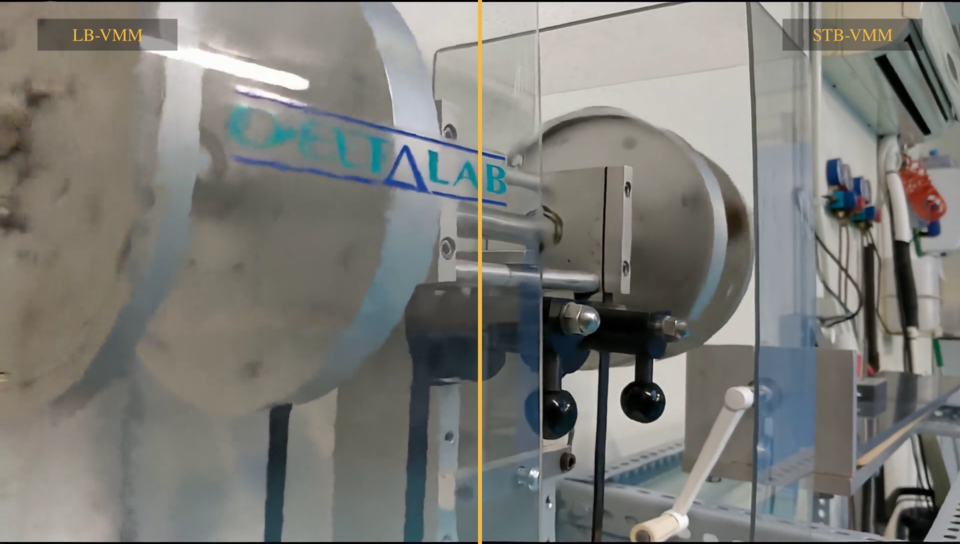}
					\caption{Split frame showing LB-VMM on the left and STB-VMM on the right}
				\end{subfigure}
				\begin{subfigure}[b]{0.2445\linewidth}
					\includegraphics[width=\linewidth]{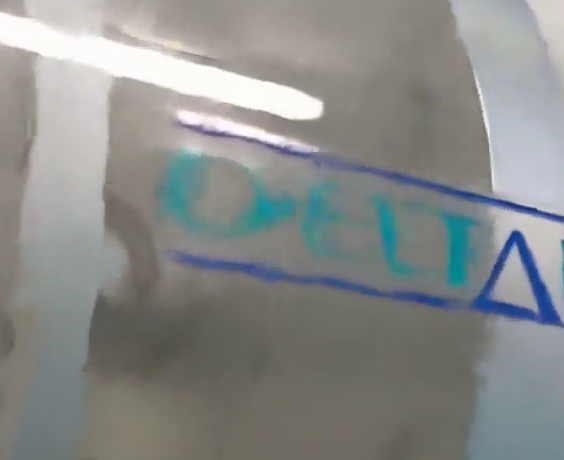}
					\caption{LB-VMM}
				\end{subfigure}
				\begin{subfigure}[b]{0.2445\linewidth}
					\includegraphics[width=\linewidth]{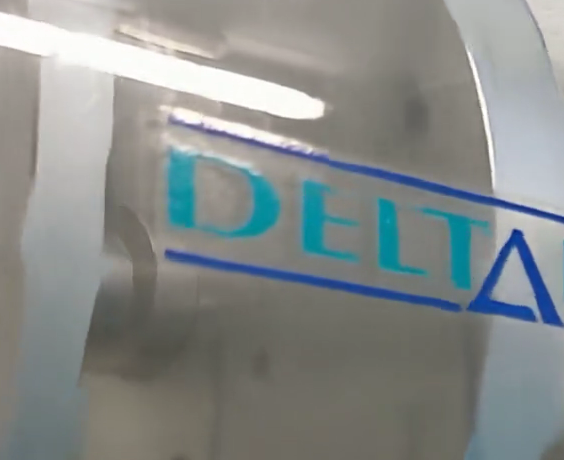}
					\caption{STB-VMM}
				\end{subfigure}
				\begin{subfigure}[b]{0.2055\linewidth}
					\includegraphics[width=\linewidth]{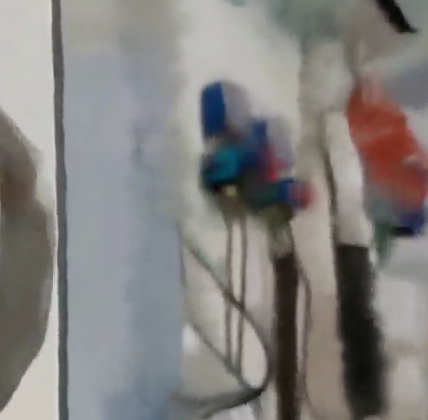}
					\caption{LB-VMM}
				\end{subfigure}
				\begin{subfigure}[b]{0.2055\linewidth}
					\includegraphics[width=\linewidth]{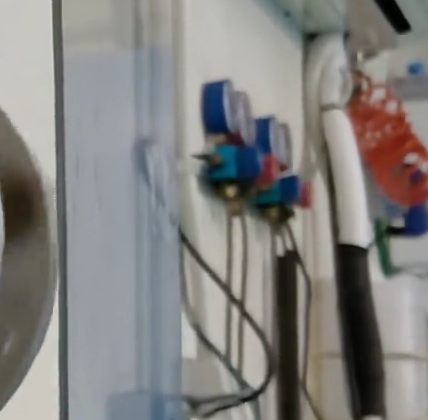}
					\caption{STB-VMM}
				\end{subfigure}
				
				\caption{Qualitative comparison of the wheel sequence. STB-VMM displays sharper letters and a better-defined background with respect to LB-VMM.}
				\label{fig:Wheel}
			\end{figure}
			
			The next example, shown in figure \ref{fig:Wheel}, was filmed in better lighting conditions, yet the quality score of the unmagnified video is no better. This might have been caused, in part, due to the framing of the sequence, which keeps only parts of the image in focus. Regardless of the base score set by the original, STB-VMM clearly outperforms LB-VMM, with better-defined letters and a much more clear background. In terms of motion magnification, both methods display good quality magnification.
			
			\begin{figure}[h]
				\centering
				\begin{subfigure}[b]{0.231\linewidth}
					\includegraphics[width=\linewidth]{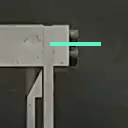}
					\caption{Slice location}
				\end{subfigure}
				\begin{subfigure}[b]{0.231\linewidth}
					\includegraphics[width=\linewidth]{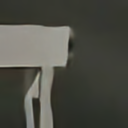}
					\caption{LB-VMM}
				\end{subfigure}
				\begin{subfigure}[b]{0.231\linewidth}
					\includegraphics[width=\linewidth]{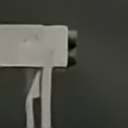}
					\caption{STB-VMM}
				\end{subfigure}
				\\
				\begin{subfigure}[b]{0.866\linewidth}
					\includegraphics[width=\linewidth]{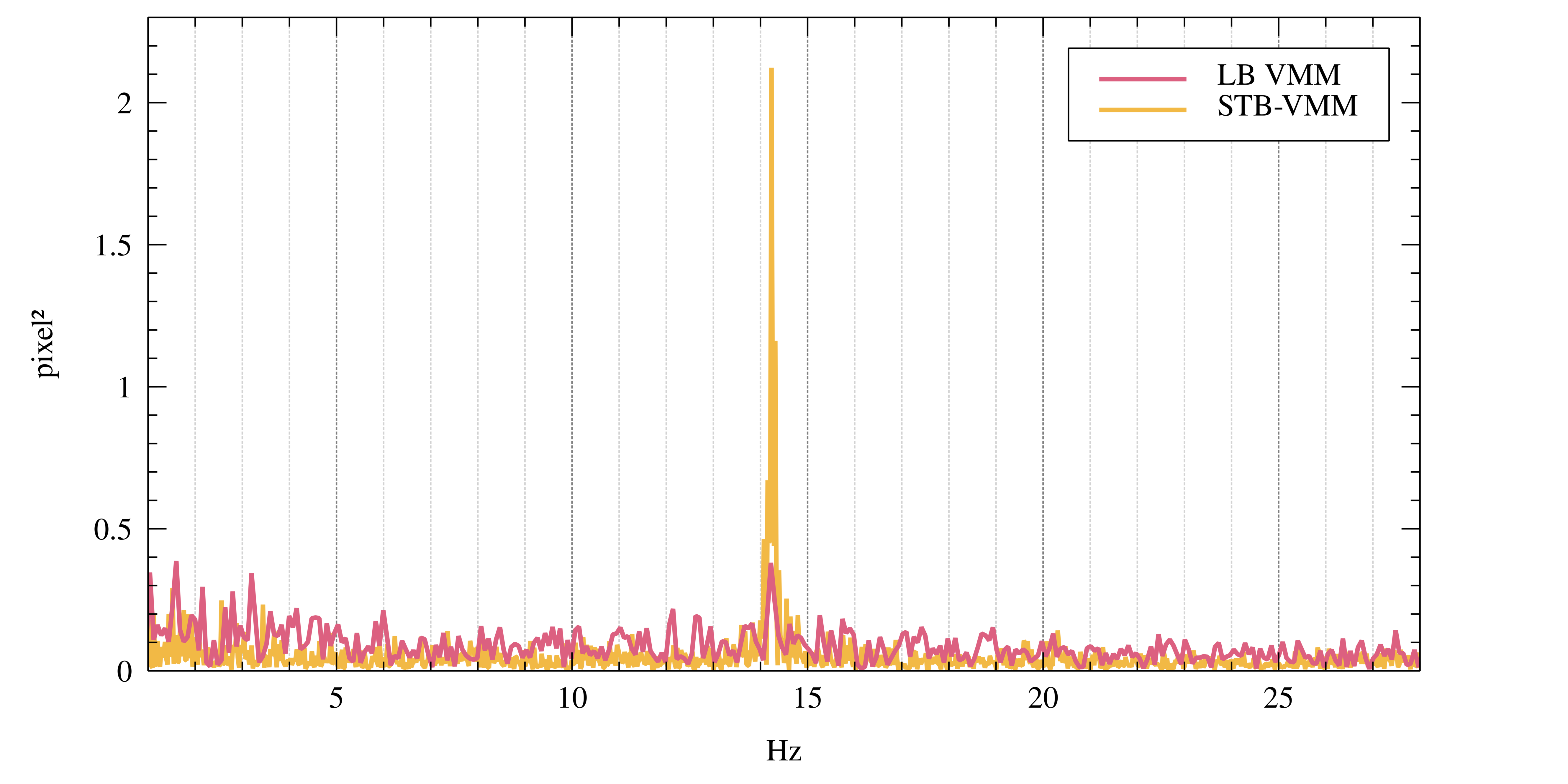}
					\caption{FFT graph comparison} 
					\label{subfig:Graph}
				\end{subfigure}
				
				\caption{Vibration readings on the Building$_{00}$ sequence. While the noise floor remains the same on both readings, the FFT obtained using STB-VMM displays a much more prominent peak at 14.25 Hz.}
				\label{fig:Signal}
			\end{figure}
			
			On the other hand, the building sequence (Building$_{00}$) is the only benchmark where LB-VMM outperforms on average STB-VMM. Nevertheless, the better edge stability offered by STB-VMM enables the authors to obtain better frequency readings from the magnified video. Such application is interesting in technical fields where vibration needs to be monitored, such as in structural health monitoring \cite{Lado2022,Font2022,Perez2021a,PEREZ2019432}. Figure \ref{fig:Signal} shows the cropped upper right corner of the building \cite{figueiredo_structural_2009} and the slice used for frequency measuring. Below, in figure \ref{subfig:Graph}, the FFTs obtained from the movement of the sequences are plotted. While both sequences detect a peak at 14.25 Hz, STB-VMM produces a much cleaner signal. During the experiment, the building was intentionally excited with an electrodynamic shaker reproducing a 14.25 Hz sine wave.
			
			The authors acknowledge that image quality can be a somewhat subjective metric and recommend watching the comparison videos attached in the supplementary materials.
		
		{\color{black}\subsection{Limitations}
			In spite of the favorable comparisons, LB-VMM still has a significant advantage in computing time over STB-VMM. With our hardware setup\footnote{AMD Ryzen 9 5950X; Nvidia RTX 3090}, LB-VMM magnifies the baby \cite{wu_eulerian_2012} sequence, consisting of 300 960x576 frames, in approximately 76 seconds. Meanwhile, STB-VMM almost doubles the compute time, clocking in at 130 seconds for the exact same sequence. Software optimizations combined with upcoming improvements in hardware might help mitigate STB-VMM's compute time shortcomings. 
		}
 
	\section{Conclusions}
		This work presents a new state-of-the-art model for video motion magnification based on the Swin Transformer that has been shown to outperform previous state-of-the-art learning-based models. The new model displays better noise tolerance characteristics, a less blurry output image, and better edge stability, resulting in clearer and less noisy magnification with very few, if any, artifacts.
		
		On the downside, the new model requires more computing resources than previous models and cannot be run in real-time like phase-based methods \cite{wadhwa_phase-based_2013}. Nevertheless, applications that require precise magnification for vibration monitoring \cite{Lado2022} could greatly benefit from improvements in the technology. Further work will address the integration of this model in specific applications that require precise vibration monitoring and could benefit from a full-field solution like a camera instead of installing and wiring multiple contact sensors such as accelerometers.
	
	\section*{Acknowledgements}
		The authors would like to gratefully acknowledge the support and funding of the Catalan Agency for Business Competitiveness (ACCIÓ) through the project INNOTEC ISAPREF 2021. Furthermore, the first author would like to acknowledge a Doctoral Scholarship from IQS. Finally, the authors would like to thank Dr. Eduardo Blanco from the University of Arizona and Dr. Ariadna Chueca de Bruijn for their help.
	
	\section*{Declaration of Competing Interest}
		The authors declare that they have no known competing financial interests or personal relationships that could appear to influence the work reported in this paper.

	\bibliographystyle{unsrt}
	\bibliography{references}
\end{document}